\documentclass[
]{ceurart}

\usepackage{todonotes}
\usepackage{tikz}
\usepackage{pgfplots}

\begin{document}

\copyrightyear{2021}
\copyrightclause{Copyright for this paper by its authors.
  Use permitted under Creative Commons License Attribution 4.0
  International (CC BY 4.0).}

\conference{CLEF 2021 -- Conference and Labs of the Evaluation Forum, 
	September 21--24, 2021, Bucharest, Romania}

\newcommand{\MacquarieUniversity}{Macquarie University}
\newcommand{\MQ}{MQ}

\title{Query-Focused Extractive Summarisation for Finding Ideal Answers to Biomedical and COVID-19 Questions}
\title[mode=alt]{\MacquarieUniversity\ Participation at BioASQ Synergy and BioASQ9b Phase~B}

\author[1,2]{Diego Moll\'a}[%
orcid=0000-0003-4973-0963,
email=Diego.Molla-Aliod@mq.edu.au,
url=https://researchers.mq.edu.au/en/persons/diego-molla-aliod,
]
\author[1]{Urvashi Khanna}[%
orcid=0000-0003-2345-5596,
email=Urvashi.Khanna@mq.edu.au,
]
\author[1]{Dima Galat}[%
email=dima.galat@gronade.com,
]
\author[2,3]{Vincent Nguyen}[%
orcid=0000-0003-1787-8090,
email=Vincent.Nguyen@anu.edu.au,
url=https://ngu.vin,
]
\author[3]{Maciej Rybinski}[%
email=maciek.rybinski@data61.csiro.au,
url=https://people.csiro.au/R/M/maciek-rybinski,
]
\address[1]{Macquarie University, Australia}
\address[2]{CSIRO Data61, Australia}
\address[3]{Australian National University, Australia}



\begin{abstract}
This paper presents \MacquarieUniversity's participation to the BioASQ Synergy Task, and BioASQ9b Phase~B. In each of these tasks, our participation focused on the use of query-focused extractive summarisation to obtain the ideal answers to medical questions. The Synergy Task is an end-to-end question answering task on COVID-19 where systems are required to return relevant documents, snippets, and answers to a given question. Given the absence of training data, we used a query-focused summarisation system that was trained with the BioASQ8b training data set and we experimented with methods to retrieve the documents and snippets. Considering the poor quality of the documents and snippets retrieved by our system, we observed reasonably good quality in the answers returned. For phase B of the BioASQ9b task, the relevant documents and snippets were already included in the test data. Our system split the snippets into candidate sentences and used BERT variants under a sentence classification setup. The system used the question and candidate sentence as input and was trained to predict the likelihood of the candidate sentence being part of the ideal answer. The runs obtained either the best or second best ROUGE-F1 results of all participants to all batches of BioASQ9b. This shows that using BERT in a classification setup is a very strong baseline for the identification of ideal answers.
\end{abstract}


\maketitle

\section{Introduction}

Supervised approaches to query-focused summarisation have the inherent problem of the paucity of annotated data.
 This problem has been highlighted, for example, by~\cite{xu2021text}, and the biomedical domain is no exception. The BioASQ Challenge provides annotated data for multiple tasks, including question answering \cite{bioasq}. While small in comparison with other data sets (the training data set for BioASQ9b contains 3,742 questions), there may be enough to train or fine-tune systems that have been pre-trained with other data sets. The problem of paucity of annotated data, however, becomes critical for urgent tasks on new domains such as question answering on biomedical papers related to COVID-19. In early 2021, BioASQ organised the Synergy task where systems are required to develop various stages of an end-to-end question answering system. In particular, given a question phrased in plain English, participating systems were expected to retrieve relevant documents from the CORD-19 collection \cite{wang-etal-2020-cord} and relevant snippets. Optionally, the systems could complete the final stage of question answering by returning exact and/or ideal answers. There was no annotated training data available for this very specific task.

This paper describes our contribution to the BioASQ Synergy task and phase~B of the BioASQ9b challenge.\footnote{Code associated with this paper is available at \url{https://github.com/dmollaaliod/bioasq-synergy-public} and  \url{https://github.com/dmollaaliod/bioasq9b-public}.} For the BioASQ Synergy task, we use a system that has been trained on the BioASQ8b training data, whereas for phase~B of the BioASQ9b challenge we explore the use of Transformer architectures. In particular, we integrate BERT variants and fine-tune them with the BioASQ9b training data.

Prior work reports the success of BERT architectures for various tasks, by simply adding a task-specific layer and fine-tuning the system~\cite{devlin-etal-2019-bert}. BERT has also been used for finding the ideal answers in BioASQ. For example, \cite{ncu-bioasq} used BERT in both an unsupervised sentence cosine similarity setup and a supervised sentence regression setup, and~\cite{mollabioasq8b} compared the use of BERT embeddings with word2vec embeddings in a setup that directly modelled the interaction between sentence embeddings of the question and the candidate sentence, and incorporated sentence position. In our participation in BioASQ9b Phase~B, we experimented with a simpler architecture compared with~\cite{mollabioasq8b}, and obtained results that were among the top participating systems\footnote{As ranked by preliminary ROUGE results provided by the BioASQ organisers at the time of writing this paper}. These good results suggest that the internal Transformer-based architecture of BERT suffices to model the interaction between the question and the candidate sentence.

This paper is structured as follows. Section~\ref{sec:synergy} describes our contribution to the Synergy task. Section~\ref{sec:bioasq9b} describes our participation in BioASQ9b. Section~\ref{sec:conclusions} summarises and concludes this paper.

\section{Synergy}\label{sec:synergy}

Our contribution to the Synergy task focused on leveraging the use of a pre-trained question answering system. In particular, we used one of the systems proposed by~\cite{mollabioasq8b}, which was trained on the BioASQ8b training data, and was designed as a classifier that identified whether a candidate sentence was part of the ideal answer. 

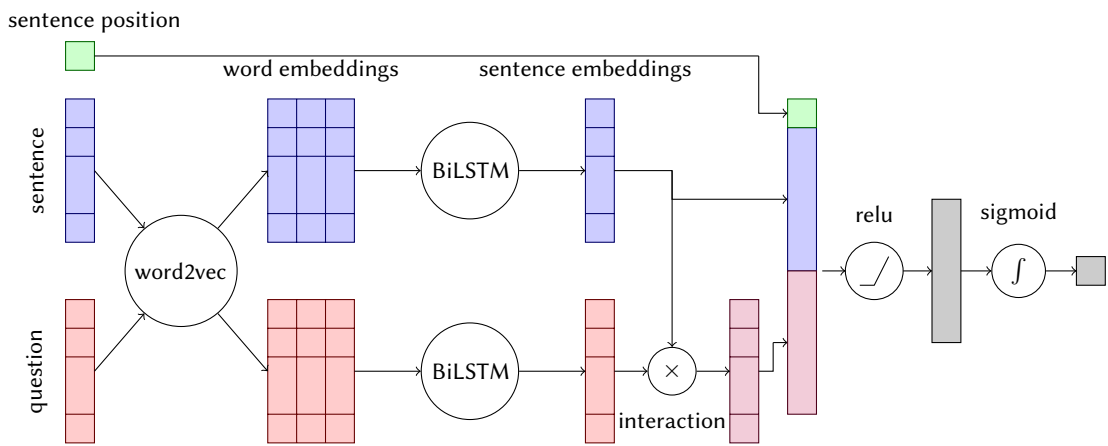
\begin{figure}
    \centering
  \centering
    \begin{tikzpicture}[scale=0.38]
    \footnotesize
    \filldraw[fill=blue!20!white, draw=blue!40!black] (0,0) rectangle (1,5) (0,1) -- (1,1) (0,3) -- (1,3) (0,4) -- (1,4);
    \filldraw[fill=red!20!white, draw=red!40!black] (0,-7) rectangle (1,-2) (0,-6) -- (1,-6) (0,-4) -- (1,-4) (0,-3) -- (1,-3);
    \draw (-1,2.5) node[rotate=90] {sentence};
    \draw (-1,-4.5) node[rotate=90] {question};

    \draw (4,-1) node [circle,draw,align=center,text width=1.2cm] (em) {word2vec};
    \draw (8.5,6) node {word embeddings};
    \filldraw[fill=blue!20!white, draw=blue!40!black] (7,0) rectangle (10,5) (7,1) -- (10,1) (7,3) -- (10,3) (7,4) -- (10,4) (8,0) -- (8,5) (9,0) -- (9,5);
    \filldraw[fill=red!20!white, draw=red!40!black] (7,-7) rectangle (10,-2) (7,-6) -- (10,-6) (7,-4) -- (10,-4) (7,-3) -- (10,-3) (8,-7) -- (8,-2) (9,-7) -- (9,-2);

    \draw[->] (1,2.5) -- (em);
    \draw[->] (1,-4.5) -- (em);

    \draw[->] (em) -- (7,2.5);
    \draw[->] (em) -- (7,-4.5);
    \draw (14,2.5) node [circle,draw,align=center,text width=1cm] (sr) {BiLSTM};
    \draw (14,-4.5) node [circle,draw,align=center,text width=1cm] (qr) {BiLSTM};
    \draw (18,6) node {sentence embeddings};
    \filldraw[fill=blue!20!white, draw=blue!40!black] (18,0) rectangle (19,5) (18,1) -- (19,1) (18,3) -- (19,3) (18,4) -- (19,4);
    \filldraw[fill=red!20!white, draw=red!40!black] (18,-7) rectangle (19,-2) (18,-6) -- (19,-6) (18,-4) -- (19,-4) (18,-3) -- (19,-3);

    \draw[->] (10,2.5) -- (sr);
    \draw[->] (sr) -- (18,2.5);
    \draw[->] (10,-4.5) -- (qr);
    \draw[->] (qr) -- (18,-4.5);

    \draw (21,-4.5) node [circle,draw] (t) {$\times$};
    \filldraw[fill=purple!20!white, draw=purple!40!black] (23,-7) rectangle (24,-2) (23,-6) -- (24,-6) (23,-4) -- (24,-4) (23,-3) -- (24,-3);

    \draw[->] (19,2.5) -| (t);
    \draw[->] (19,-4.5) -- (t);

    \draw[->] (21,2.5) |- (25,1.5);
    \draw[->] (t) -- (23,-4.5);
    \draw (24,-4.5) -| (24.5,-3.5);
    \draw[->] (24.5,-3.5) -- (25,-3.5);

    \draw (21,-6.2) node {interaction};

    \filldraw[fill=blue!20!white, draw=blue!40!black] (25,-1) rectangle (26,4);
    \filldraw[fill=purple!20!white, draw=purple!40!black] (25,-1) rectangle (26,-6);
    \draw (28,-1) circle[radius=1] (27.5,-1.5) -- (28,-1.5) -- (28.5,-0.5);
    \draw (28,1) node {relu};

    \filldraw[fill=black!20!white, draw=black] (30,-3.5) rectangle (31,1.5);

    \draw[->] (26.2,-1) -- (27,-1);
    \draw[->] (29,-1) -- (30,-1);

    \draw (33,-1) node[circle,draw] (sig) {$\int$};
    \filldraw[fill=black!20!white, draw=black] (35,-1.5) rectangle (36,-0.5);
    \draw (33,1) node[text width=1cm] {sigmoid};

    \draw[->] (31,-1) -- (sig);
    \draw[->] (sig) -- (35,-1);
    
    \filldraw[fill=green!20!white, draw=green!40!black] (0,6) rectangle (1,7);
    \draw (1,6.5) -- (24,6.5);
    \draw[->] (24,6.5) |- (25,4.5);
    \filldraw[fill=green!20!white, draw=green!40!black] (25,5) rectangle (26,4);
    \draw (1,7.7) node {sentence position};
  \end{tikzpicture}
    \caption{Architecture of the question answering system used for the Synergy task.}
    \label{fig:bioasq8b}
\end{figure}

Figure~\ref{fig:bioasq8b} shows the architecture of the question answering system.
This corresponds to the system referred to as ``NNC'' by~\cite{mollabioasq8b}. The input consists of a question, a candidate sentence, and the candidate sentence position. The system uses Word2Vec trained on PubMed data to obtain the word embeddings of the question and the sentence. These word embeddings are converted to sentence embeddings through a layer of bi-directional LSTM chains. The interaction between the question and the sentence embeddings is modelled by applying element-wise multiplication. The result of this multiplication is concatenated to the sentence embeddings and the sentence positions. There is an intermediate hidden layer with dropout, followed by the final classification layer. The loss function is binary cross-entropy, and the target labels (0 or 1) were generated based on the ROUGE-F1 score of the candidate sentence with respect to the corresponding ideal answer.

The hyperparameters of the system are: Number of epochs=10; batch size=1024; dropout=0.7; hidden layer size = 50; embeddings size=100; sentence length clipped to 300 tokens.

The following sequence of steps was used to generate the candidate sentences that were fed as input to the question answering system:

\begin{enumerate}
    \item Obtain the list of candidate documents, sorted by relevance. For this, we used the search API provided by the organisers of the BioASQ Synergy task. We also experimented with the use of sentence BERT fine-tuned with the BioASQ data as described in Section~\ref{sec:docs}.
    \item Split the documents into sentences and select and rank the most relevant sentences. For this, we experimented with various methods described in Section~\ref{sec:snippets}. The resulting sentences were used as candidate sentences to be processed by the question answering system.
\end{enumerate}

\subsection{Document Retrieval}
\label{sec:docs}

We experimented with three different approaches to document retrieval. These are listed below and named for further reference in this paper.

\paragraph{DocAPI}
Most of our runs used the search API provided by the organisers of the BioASQ Synergy task. In preliminary experiments, we observed that the default results returned by the API were correlated with the cosine similarity with the question. We therefore concluded that the API returned the results ranked by some sort of relevance. Consequently, we selected the top $n$ documents, where $n$ depended on the round number (50 for round 1, and 100 for every subsequent round).

\paragraph{DocNIR(untuned)}
 We submitted one run based on the Neural Index Retrieval (NIR) methodology outlined in~\cite{nguyen-etal-2020-pandemic}. This document retrieval method combines a traditional inverted index with a neural index of the document collection. Specifically, the document relevance scores for each query are obtained by interpolating the normalised BM25 scores (so, the relevance score based on the use of a traditional inverted index) with a cosine similarity score between the neural representations of the query and the document. The neural representations are obtained via a sBERT~\cite{reimers-2019-sentence-bert} model pre-trained on the target corpus and fine-tuned on a natural language inference task.\footnote{We used the ``manueltonneau/clinicalcovid-bert-nli'' model from the huggingface transformers repository.}
 
\paragraph{DocNIR(tuned)}
  In round 4, we also experimented with document retrieval based on the use of sBERT~\cite{reimers-2019-sentence-bert} fine-tuned with the BioASQ data. The retrieval model is, in essence, similar to the NIR method outlined above. The main difference is that the sBERT model 
  is additionally fine-tuned on the target task training data (specifically, the relevance feedback available from the previous rounds). Another notable difference is that we used the neural component only to re-score (still using the BM25 for interpolation) the top-200 documents retrieved by the BM25 model for each query (making it, effectively, a re-ranker).

The final runs used the top 10 documents, after removing those that were in previous feedback.

\subsection{Snippet Retrieval}
\label{sec:snippets}

We experimented with several approaches to identify and rank the relevant snippets as described below.

\paragraph{SnipCosine}
Our baseline snippet retrieval system was based on the tf.idf cosine similarity between the question and the input document sentences. In particular, each document retrieved by the document retrieval system (after removing false positives as indicated by the feedback form previous rounds) was split into sentences (using NLTK's sentence tokeniser). Then, for each document, the top 3 sentences were extracted. To identify the top 3 sentences, we used cosine similarity between the tf.idf vector of the question and the candidate sentence. For each document, the top 3 sentences were ranked by order of occurrence in the document (not by order of similarity). These sentences were then collated by order of document relevance.

\paragraph{SnipQA}
Our second approach used the BioASQ8b question answering system (Figure~\ref{fig:bioasq8b}) to rank the document input sentences. The rationale for this approach was that the BioASQ8b question answering system had been trained to score sentences based on their likelihood of being part of the ideal answer, and we wanted to know whether such a system could be used, without fine-tuning, as a snippet re-ranker. As with the baseline system, the documents were split into sentences. These sentences were then scored using the question-answering system, and the top 3 sentences per document were selected and ranked by order of occurrence. The resulting sentences were then collated by order of document relevance.

\paragraph{SnipSBERT}
A third approach was based on the use of sBERT \cite{reimers-2019-sentence-bert}, trained for passage retrieval. Using the full CORD-19\footnote{https://www.semanticscholar.org/cord19} dataset we have tried to retrieve the most relevant snippets by minimising a cosine distance between a question and a sentence in the dataset. Two variants were implemented: SnipSBERT(a) used the output of the Synergy API and returned the top 3 snippets, whereas SnipSBERT(b) searched the CORD-19 data directly and returned the top 3 or top 5 snippets.

The final runs used the top 10 snippets, after removing those that were in previous feedback.

\subsection{Answer Generation}\label{sec:qa}

In all of our experiments, answer generation used the same process as illustrated in Figure~\ref{fig:bioasq8b} to conduct query-focused extractive summarisation. In particular, as in the original paper \cite{mollabioasq8b}, given a question, sentence, and sentence position, the system predicted the probability that the sentence has high ROUGE-F1 score with the ideal answer. We obtained the relevant sentences using the methods for snippet retrieval detailed in Section~\ref{sec:snippets}. Then, irrelevant sentences (as indicated by feedback from previous rounds) were removed. The position of the remaining sentences was indicated by their order after the snippet retrieval stage and after removing irrelevant sentences. With this information, the question answering system returned the sentence score. The answer was obtained by selecting the top $n$ sentences, and sorting them by order of appearance in the list of snippets. The value of $n$ depended on the question type as listed in Table~\ref{tab:valueofn}.
\begin{table}
    \centering
    \caption{Number of sentences selected, for each question type}
    \label{tab:valueofn}
    \begin{tabular}{ccccc}
    & \textbf{Summary} & \textbf{Factoid} & \textbf{Yesno} & \textbf{List} \\
    \midrule
    \textbf{n}     &  6 & 2 & 2 & 3\\
    \end{tabular}
\end{table}

\subsection{Results of the Synergy Task}\label{sec:synergyruns}

All runs submitted to the Synergy Task use the same approach to generate the ideal answers (Section~\ref{sec:qa}) and we experimented with combinations of the approaches to retrieve the documents (Section~\ref{sec:docs}) and the snippets (Section~\ref{sec:snippets}). The specific set up of each run, and the results, are detailed in
Tables \ref{tab:synergydocs}, \ref{tab:synergysnips}, and~\ref{tab:synergyqa}.

\begin{table}
    \centering
    \caption{Document retrieval results of the submission to Synergy. Metric: F1. Legend: DocAPI$^1$; DocNIR(untuned)$^2$; DocNIR(tuned)$^3$. }
    \label{tab:synergydocs}
    \begin{tabular}{lrrrr}
    Run & Round 1 & Round 2 & Round 3 & Round 4\\
    \midrule
    Best&0.3457&0.3237&0.2628&0.2375\\
    Median&0.2474&0.2387&0.1810&0.1839\\
    Worst&0.0802&0.0560&0.0179&0.0168\\
    \midrule
    \MQ-1&0.2474$^1$&0.1654$^1$&0.0973$^1$&0.1053$^1$\\
    \MQ-2&0.2474$^1$&0.1654$^1$&0.0973$^1$&0.1053$^1$\\
    \MQ-3&&0.1654$^1$&0.0973$^1$&0.1053$^1$\\
    \MQ-4&&0.1654$^1$&&0.1510$^2$\\
    \MQ-5&&&&0.1762$^3$\\
  \end{tabular}
\end{table}

\begin{table}
    \centering
    \caption{Snippet retrieval results of the submission to Synergy. Metric:F1. Legend: DocAPI$\rightarrow$SnipCosine$^1$; DocAPI$\rightarrow$SnipQA$^2$; DocNIR(untuned)$\rightarrow$SnipCosine$^3$; DocNIR(tuned)$\rightarrow$SnipCosine$^4$.}
    \label{tab:synergysnips}
\begin{tabular}{lrrrr}
    Run & Round 1 & Round 2 & Round 3 & Round 4\\
    \midrule
    Best&0.2712&0.1885&0.2026&0.1909\\
    Median&0.2021&0.1634&0.1645&0.1461\\
    Worst&0.0396&0.0204&0.0037&0.0078\\
    \midrule
    \MQ-1&0.1414$^1$&0.0704$^1$&0.0462$^1$&0.0640$^1$\\
    \MQ-2&0.1380$^2$&0.0706$^2$&0.0462$^2$&0.0657$^2$\\
    \MQ-3&&0.0709$^2$&0.0473$^2$&0.0634$^2$\\
    \MQ-4&&0.0695$^2$&&0.0798$^3$\\
    \MQ-5&&&&0.0912$^4$\\
  \end{tabular}
\end{table}
  
\begin{table}
    \centering
    \caption{Ideal answer results of the submission to Synergy. Metrics: ROUGE-SU F1 $|$ Average human evaluation. Legend: DocAPI$\rightarrow$SnipCosine$\rightarrow$QA$^1$; DocAPI$\rightarrow$SnipQA$\rightarrow$QA$^2$; DocNIR(untuned)$\rightarrow$SnipCosine$\rightarrow$QA$^3$; DocNIR(tuned)$\rightarrow$SnipCosine$\rightarrow$QA$^4$;
    SnipSBERT(a)$\rightarrow$QA$^5$;
    SnipSBERT(b)$\rightarrow$QA$^6$.}
    \label{tab:synergyqa}
  \begin{tabular}{lrrrr}
    Run & Round 1 & Round 2 & Round 3 & Round 4\\
    \midrule
    Best&&0.0749 $|$ 3.672&0.1170 $|$ 4.185&0.1254 $|$ 3.662\\
    Median&&0.0565 $|$ 3.127&0.0883 $|$ 3.517&0.0857 $|$ 3.157\\
    Worst&&0.0096 $|$ 0.667&0.0181 $|$ 0.750&0.0221 $|$ 0.705\\
    \midrule
    \MQ-1&&0.0567$^1$ $|$ 3.015&0.0883$^1$ $|$ 3.517&0.0971$^1$ $|$ 3.140\\
    \MQ-2&&0.0565$^2$ $|$ 2.965&0.0926$^2$ $|$ 3.542&0.0912$^2$ $|$ 3.157\\
    \MQ-3&&0.0436$^5$ $|$ 2.670&0.0467$^6$ $|$ 3.062&0.0515$^6$ $|$ 2.982\\
    \MQ-4&&0.0500$^6$ $|$ 3.047&&0.0857$^3$ $|$ 3.190\\
    \MQ-5&&&&0.0757$^4$ $|$ 3.060\\
  \end{tabular}
\end{table}

In document retrieval (Table~\ref{tab:synergydocs}), the NIR retrieval approaches outperformed the baseline that used the API provided by the BioASQ organisers. Also, we observed best results when the document retrieval system was tuned with the BioASQ data. Having said this, compared with the other submissions to the Synergy tasks, the document retrieval systems performed poorly, especially on rounds 2 to 4.

In snippet retrieval, some runs used the output of DocAPI, others used the output of DocNIR(untuned and tuned). We experimented with snippet re-ranking using SnipCosine and SnipQA as detailed in Table~\ref{tab:synergysnips}. None of the runs used SnipSBERT because of problems meeting the format requirements.\footnote{The Synergy task requires all snippets to include the character offsets. However, our implementation did not provide this information.} We observed variability of results in the runs that used the sequence DocAPI$\rightarrow$SnipQA due to the undeterministic nature of the question answering module. Overall, all results were very similar, and comparatively worse than the results of other runs. In fact, our runs were near the bottom of the leaderboard in rounds 2 to 4.

In ideal answer generation, the input to the question answering module used a sequence of document retrieval followed by snippet retrieval as detailed in Table~\ref{tab:synergyqa}. We also included runs that used SnipSBERT for snippet retrieval. Considering the poor results of the snippet retrieval stage, the ideal answer results were relatively good and they were approximately around the median of all submissions. This gives some indication that the question answering system, trained on medical data but not on data containing COVID-19, was relatively robust. Even though the absolute values of the ROUGE-S1 F1 scores were rather low, the average scores of the human evaluation of most of our runs were above 3 in a scale from 0 to 4.

\section{BioASQ9b Phase B}\label{sec:bioasq9b}

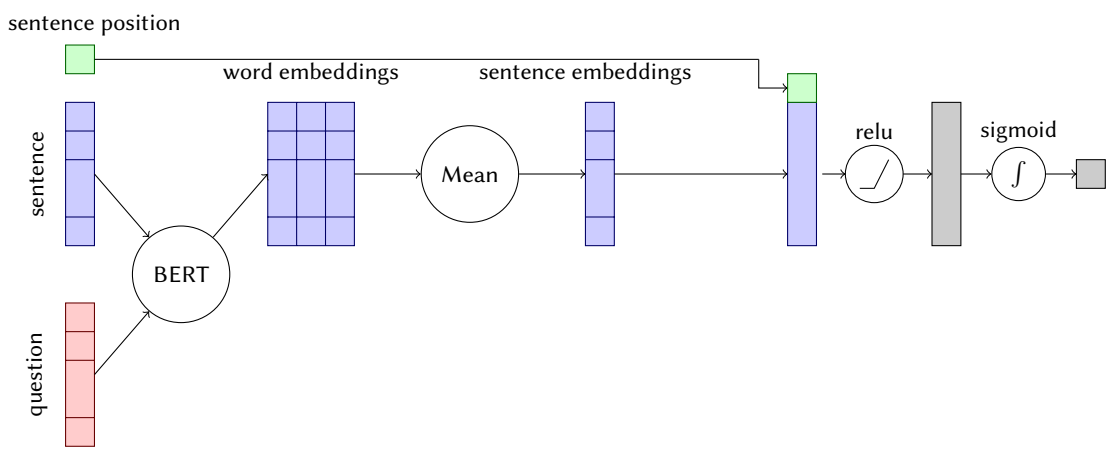
\begin{figure}
    \centering
    \begin{tikzpicture}[scale=0.38]
    \footnotesize
    \filldraw[fill=blue!20!white, draw=blue!40!black] (0,0) rectangle (1,5) (0,1) -- (1,1) (0,3) -- (1,3) (0,4) -- (1,4);
    \filldraw[fill=red!20!white, draw=red!40!black] (0,-7) rectangle (1,-2) (0,-6) -- (1,-6) (0,-4) -- (1,-4) (0,-3) -- (1,-3);
    \draw (-1,2.5) node[rotate=90] {sentence};
    \draw (-1,-4.5) node[rotate=90] {question};

    \draw (4,-1) node [circle,draw,align=center,text width=1cm] (em) {BERT};
    \draw (8.5,6) node {word embeddings};
    \filldraw[fill=blue!20!white, draw=blue!40!black] (7,0) rectangle (10,5) (7,1) -- (10,1) (7,3) -- (10,3) (7,4) -- (10,4) (8,0) -- (8,5) (9,0) -- (9,5);

    \draw[->] (1,2.5) -- (em);
    \draw[->] (1,-4.5) -- (em);

    \draw[->] (em) -- (7,2.5);
    \draw (14,2.5) node [circle,draw,align=center,text width=1cm] (sr) {Mean};
    \draw (18,6) node {sentence embeddings};
    \filldraw[fill=blue!20!white, draw=blue!40!black] (18,0) rectangle (19,5) (18,1) -- (19,1) (18,3) -- (19,3) (18,4) -- (19,4);

    \draw[->] (10,2.5) -- (sr);
    \draw[->] (sr) -- (18,2.5);





    \draw[->] (19,2.5) -- (25,2.5);

    \filldraw[fill=blue!20!white, draw=blue!40!black] (25,0) rectangle (26,5);
    \draw (28,2.5) circle[radius=1] (27.5,2) -- (28,2) -- (28.5,3);
    \draw (28,4) node {relu};

    \filldraw[fill=black!20!white, draw=black] (30,0) rectangle (31,5);

    \draw[->] (26.2,2.5) -- (27,2.5);
    \draw[->] (29,2.5) -- (30,2.5);

     \draw (33,2.5) node[circle,draw] (sig) {$\int$};
\filldraw[fill=black!20!white, draw=black] (35,2) rectangle (36,3);
    \draw (33,4) node[text width=1cm] {sigmoid};

    \draw[->] (31,2.5) -- (sig);
    \draw[->] (sig) -- (35,2.5);
    
    \filldraw[fill=green!20!white, draw=green!40!black] (0,6) rectangle (1,7);
    \draw (1,6.5) -- (24,6.5);
    \draw[->] (24,6.5) |- (25,5.5);
    \filldraw[fill=green!20!white, draw=green!40!black] (25,6) rectangle (26,5);
    \draw (1,7.7) node {sentence position};
  \end{tikzpicture}

    \caption{Architecture of the question answering system used for BioASQ 9b, Phase B.}
    \label{fig:bioasq9b}
\end{figure}

The system that participated in BioASQ9b Phase B focused on the use of BERT-based architectures for query-focused extractive summarisation. The experiments reported by \cite{mollabioasq8b} indicated that replacing Word2Vec with BERT in the system of Figure~\ref{fig:bioasq8b} only gave a minor improvement of the results. Subsequent (unpublished) experiments also appeared to indicate that using BERT as an end-to-end system, without adding the multiplication layer between question and sentence, plus the addition of the sentence position for the final classification layer, leads to similar or better results. This motivated us to experiment with the use of BERT in the architecture shown in Figure~\ref{fig:bioasq9b}.
The new architecture is a simplification to that of Figure~\ref{fig:bioasq8b}, where most of the computation, including determining the interaction between the question and the sentence, is carried out by BERT. We experimented with several BERT variants as described in Section~\ref{sec:bertvariants}. As with the system of Figure~\ref{fig:bioasq8b} and the system by \cite{mollabioasq8b}, the system performs extractive summarisation and it is trained to predict whether the candidate sentence has a high ROUGE-SU4 F1 score with the ideal answer. In particular, the label of the training set was~1 if the sentence was among the 5 sentences with highest ROUGE-SU4 F1 score, and~0 otherwise. The final ideal answer is obtained by selecting the top $n$ sentences, and these sentences are presented in order of appearance in the input snippets. The value of $n$ is as shown in Table~\ref{tab:valueofn}.

The question and sentence were fed to BERT in the standard approach defined by the creators of BERT \cite{devlin-etal-2019-bert}. In particular, the question and sentence were input as two separate text segments in the following order: first the ``[CLS]'' special token, then the question, then the sentence separator ``[SEP]'', and finally the candidate sentence.

Instead of passing the embedding of the ``[CLS]'' special token to the classification layer, we decided to use the embeddings of the tokens forming the candidate sentence. These embeddings were mean pooled in order to obtain the sentence embeddings.

\subsection{BERT Variants}\label{sec:bertvariants}

We experimented with the following BERT variants. All of these variants were based on models made available by the Huggingface transformers repository\footnote{https://huggingface.co/transformers/}.

\paragraph{BERT} We used huggingface's model ``bert-base-uncased''.

\paragraph{BioBERT} Given the medical nature of BioASQ, we tried BioBERT, which uses the same architecture as BERT base, and has been fine-tuned with PubMed articles \cite{biobert}. We used huggingface's model ``monologg/biobert\_v1.1\_pubmed''.

\paragraph{DistilBERT} DistilBERT's architecture is a reduced version of BERT, which has been trained to replicate the soft predictions made by BERT \cite{distilbert}. The resulting system is faster to train, and reportedly nearly as accurate as BERT. We used huggingface's model ``distilbert-base-uncased''.

\paragraph{ALBERT} ALBERT uses parameter reduction techniques that allow faster training and with lower memory consumption. This enables the use of larger numbers of transformer layers and larger embedding sizes \cite{lan2019albert}. We used huggingface's model ``albert-xxlarge-v2''.

\paragraph{ALBERT-SQuAD} This variant of ALBERT has been fine-tuned with data from SQuAD, a well-known data set for question answering systems in the context of reading comprehension \cite{rajpurkar2018know}. We used huggingface's model ``mfeb/albert-xxlarge-v2-squad2''.

\paragraph{ALBERT-QA}
This final variant of ALBERT was obtained using ALBERT-SQUAD as a starting point (using huggingface's model ``mfeb/albert-xxlarge-v2-squad2''). Then, the model was fine-tuned by adding a SQuAD-style question answering classification layer and trained on the BioASQ training set, using the exact answers as labels. For this fine-tuning stage, only factoid questions were used. The system that implemented this fine-tuning is one of the systems described by \cite{UrvashiBioASQ9b}.


\vspace{0.5cm}

In all of our experiments, we froze all BERT layers and only trained the hidden and classification layers. The reason for this decision was that, in preliminary experiments with unfrozen BERT layers, we observed the catastrophic forgetting effect where all the pre-trained information was lost, and decided to leave the study of fine-tuning strategies of the BERT layers for further work.

Table~\ref{tab:cv} shows the results of 10-fold cross-validation on the BioASQ9b training data. The table also shows the values of the differing hyperparameters of the best systems as found through grid search. The hyperparameters common to all systems were: batch size=32; hidden layer size=50; sentence length clipped to 250 tokens. 
\begin{table}
    \centering
    \caption{Results of 10-fold cross-validation using the BioASQ9b training data. Metric: ROUGE SU4 F1.}
    \label{tab:cv}
\begin{tabular}{lrrrrr}
    &\multicolumn{2}{c}{Number of Parameters}\\
    System & Full & Trained & Epochs & Dropout & SU4-F1\\
    \midrule
    BERT&109,520,791&38,551&8&0.8&0.2779\\
    BioBERT&108,348,823&38,551&1&0.7&0.2798\\
    DistilBERT&66,401,431&38,551&1&0.6&0.2761\\
    ALBERT&222,800,535&204,951&5&0.5&0.2866\\
    ALBERT-SQuAD&222,800,535&204,951&5&0.7&0.2846\\
    ALBERT-QA&222,800,535&204,951&5&0.4&\textbf{0.2875}\\
  \end{tabular}
\end{table}
Overall, all results are similar, but we can observe that BioBERT outperforms BERT, in line with most prior work (but in contrast with \cite{mollabioasq8b}). We can also observe an improvement of the results of the three ALBERT variants. This is possibly due to the larger architecture sizes. The fact that ALBERT-QA has a slightly better result than the other ALBERT variants is encouraging.

\subsection{Submission Results to BioASQ 9b Phase B}

The runs submitted to BioASQ9b Phase B used all the BERT variants described in Section~\ref{sec:bertvariants} except ALBERT-SQuAD. 

The preliminary evaluation results, as reported in the BioASQ website, are shown in Table~\ref{tab:bioasq9b}.\footnote{Note that the results reported in the BioASQ website (http://bioasq.org) may change in the future after the test data is enriched with further annotations.}
\begin{table}
    \centering
    \caption{Preliminary results of the submissions to BioASQ9b, Phase B.}
    \label{tab:bioasq9b}
   \begin{tabular}{llrrrrr}
     && \multicolumn{5}{c}{ROUGE-SU4}\\
     Run&System & Batch 1 & Batch 2 & Batch 3 & Batch 4 & Batch 5\\
     \midrule
     Best&&0.3410&0.3974&0.3266&0.4402&0.3893\\
     Median&&0.2536&0.1990&0.2647&0.3388&0.2666\\
     Worst&&0.1154&0.1186&0.1017&0.0886&0.1331\\
     \midrule
     \MQ-1&BERT&0.3032&0.3560&0.3057&0.3585&0.3511\\
     \MQ-2&BioBERT&0.3103&0.3615&0.3265&0.3612&\bf 0.3733\\
     \MQ-3&DistilBERT&0.3007&\bf 0.3753&0.3204&\bf 0.3681&0.3711\\
     \MQ-4&ALBERT&\bf 0.3205&0.3676&0.3100&0.3560&0.3570\\
     \MQ-5&ALBERT-QA&&0.3610&\bf 0.3266&0.3559&0.3589\\
   \end{tabular}
\end{table}
For each batch, our runs ranked among the top participating systems. In fact, ALBERT-QA was the top run of batch 3. This demonstrates that a straightforward use of BERT is a very strong baseline. As expected, BioBERT outperformed BERT. The experiments with ALBERT and ALBERT-QA in batches~4 and~5, however, were not as good as expected given our cross-validation results.

\section{Summary and Conclusions}\label{sec:conclusions}

We have presented \MacquarieUniversity's contribution to the BioASQ Synergy task and BioASQ9b Phase B (Ideal Answers).

For the synergy task, we have experimented with a question answering module that was designed for, and trained with, the data from BioASQ8b. Due to the need to produce an end-to-end system, we tried various baseline document and snippet retrieval systems. Overall, despite the poor general quality of the document and snippet retrieval systems, the results of our submissions indicate that the question answering component can generalise well to questions related to COVID-19. Further work will focus on improving the quality of the document and snippet retrieval components.

The synergy task was organised in multiple rounds such that feedback from previous rounds was available for subsequent rounds in some questions. Our system incorporated this feedback only in a trivial manner, simply by removing documents or snippets that were identified as known negatives. There has been research on relevance feedback since at least 1971 \cite{Roccio}, which could be incorporated into the system. More recent approaches, such as using a twin neural network with a contrastive loss \cite{contrastive}, may work here.

The contribution to BioASQ9b Phase B focused on the use of BERT variants within a query-focused extractive summarisation setting. The architecture concatenates the question and candidate sentence as two separate text segments, very much as is done in question-answering approaches with BERT, and the system is trained as a sentence classification system. We observe that such a simple architecture is a very strong baseline. Further work will focus on exploring further variants of BERT, and on enhancing the pre-training and fine-tuning stages.

\begin{acknowledgments}
This research was undertaken with the assistance of resources and services from the National Computational Infrastructure (NCI), which is supported by the Australian Government.

Research by Vincent Nguyen is supported by the Australian Research Training Program and the CSIRO Postgraduate Scholarship.
\end{acknowledgments}

\bibliography{bioasq}


\end{document}